\documentclass[conference,a4paper]{APSIPA2025}
\usepackage{amsmath}
\usepackage{graphicx}
\usepackage{multirow}
\usepackage{threeparttable}
\usepackage{amsmath}
\usepackage{algorithm}
\usepackage{algpseudocode}
\usepackage{booktabs}
\usepackage{amssymb}
\usepackage{pifont}
\newcommand{\cmark}{\ding{51}}%
\newcommand{\xmark}{\ding{55}}%
\usepackage{geometry}
\usepackage{float}
\usepackage{fancyhdr}
\usepackage{adjustbox}
\usepackage{url}
\usepackage{comment}

\fancypagestyle{firststyle}{
  \fancyhf{}
  \fancyhead[C]{2025 Asia Pacific Signal and Information Processing Association Annual Summit and Conference (APSIPA ASC)}
}
\usepackage{lipsum}

\newcommand\blfootnote[1]{%
  \begingroup
  \renewcommand\thefootnote{}\footnote{#1}%
  \addtocounter{footnote}{-1}%
  \endgroup
}

\begin{document}

\title{Exploring Machine Learning and Language Models for Multimodal Depression Detection}

\author{
\authorblockN{
Javier Si Zhao Hong\authorrefmark{1},
Timothy Zoe Delaya\authorrefmark{1},
Sherwyn Chan Yin Kit\authorrefmark{1},
Pai Chet Ng\authorrefmark{1},
Xiaoxiao Miao\authorrefmark{2}
}
\authorblockA{
\authorrefmark{1} Singapore Institute of Technology, Singapore \\
E-mail: \{2302655,2302663,2302669\}@sit.singaporetech.edu.sg, paichet.ng@singaporetech.edu.sg
}
\authorblockA{
\authorrefmark{2} Duke Kunshan University, China \\
E-mail: xiaoxiao.miao@dukekunshan.edu.cn
}
}

\maketitle
\thispagestyle{firststyle}
\pagestyle{fancy}

\begin{abstract}
 This paper presents our approach to the first Multimodal Personality-Aware Depression Detection Challenge, focusing on multimodal depression detection using machine learning and deep learning models. We explore and compare the performance of XGBoost, transformer-based architectures, and large language models (LLMs) on audio, video, and text features. Our results highlight the strengths and limitations of each type of model in capturing depression-related signals across modalities, offering insights into effective multimodal representation strategies for mental health prediction.\blfootnote{Xiaoxiao Miao is the corresponding author and this work was conducted while she was at SIT.}
 
\end{abstract}

\section{Introduction}

The World Health Organization recently reported that depression affects 3.8\% of the global population and 63.6\% of these cases remain undiagnosed  \cite{who2023depression}, partly due to the limited availability of healthcare services and, in some cases, financial constraints that prevent many individuals from accessing necessary medical care \cite{faisalcury2022depression}. Traditional methods, which rely primarily on self-reported questionnaires such as the PHQ-9 \cite{kroenke2001phq9} and the BDI-II \cite{beck1996bdi2ndmanual}, are limited in their ability to capture the dynamic and multifaceted nature of depressive symptoms. These methods are also prone to reporting biases and may fail to detect early or subtle changes in depressive states.

In response to these challenges, the computing community has been instrumental in advancing automatic depression detection by leveraging multimodal data \cite{dibekliouglu2017dynamic, yoon2022d, shen2022automatic, zou2022semi, cai2022multi, miranda2018amigos, koelstra2011deap, klein2011personality, lo2017genome}. 
Most recently, the first Multimodal Personality-Aware Depression Detection (MPDD) Challenge \cite{fu2025mpdd} introduced a richly annotated novel dataset that includes audio and visual recordings of participants engaging in a variety of real-world scenarios. The MPDD dataset is annotated using the PHQ-9 scale, Big Five personality traits \cite{rammstedt2007measuring}, and detailed demographic information. Compared to existing corpora, the MPDD dataset offers greater contextual diversity and annotation depth, enabling more inclusive and fine-grained modeling.

In terms of methodological approaches, current research increasingly integrates audio, visual, and textual biomarkers through progressively sophisticated computational paradigms.
Given that depression detection is a form of fine-grained, subtle emotion recognition, many researchers have drawn inspiration from emotion classification methods.
Traditional machine learning approaches typically rely on handcrafted feature extraction pipelines. For example, OpenSMILE-derived acoustic features \cite{eyben2013opensmile} are often combined with facial expression metrics extracted using computer vision tools, such as Action Units from OpenFace \cite{baltrusaitis2018openface} or emotion probabilities from facial emotion recognition models \cite{shenk2019fer}. These features are typically fed into ensemble classifiers, such as XGBoost \cite{chen2016xgboost}, or kernel-based methods, like support vector machine \cite{hearst1998support}. Some studies further enhance performance using feature selection or principal component analysis (PCA)-based dimensionality reduction prior to classification \cite{gideon2023detecting}.

More advanced systems employ hybrid architectures with modality-specific processing pipelines. A common configuration involves using separate convolutional neural networks (CNNs) for visual frames, long short-term memory (LSTM) networks for audio spectrograms, and transformer networks for textual inputs \cite{zhang2021first}. Fusion strategies vary, ranging from early fusion of low-level features to late fusion of modality-specific predictions \cite{pmc2023fusion}. Recent work has also explored cross-modal attention mechanisms using transformers to learn joint representations \cite{chen2024cross}.
Further progress has been achieved through unified architectures that combine self-supervised audio encoders for paralinguistic feature extraction, vision transformers for modeling spatio-temporal facial dynamics, and specialized language models for clinical text analysis \cite{ji2022mentalbert}. 
In recent years, large language models (LLMs) have begun to reshape the field. Multimodal emotion recognition systems such as Emotion-LLaMA \cite{NEURIPS2024_c7f43ada} integrate audio, visual, and textual inputs through emotion-specific encoders. By aligning these features within a shared latent space and applying a modified LLaMA model with instruction tuning, Emotion-LLaMA significantly enhances both emotional recognition and reasoning capabilities.

Among these approaches, XGBoost, transformer-based models, and LLMs have each achieved state-of-the-art results at different stages of research.
However, the MPDD Challenge launched in 2025 introduces new complexities that may affect model performance when applying different methods.
Inspired by this, the present study systematically evaluates and compares the effectiveness of three representative model classes, XGBoost, transformer-based models, and LLMs, on the MPDD dataset. Following the official challenge protocol, we assess the strengths and limitations of each model across modalities, with the goal of identifying their respective potentials for advancing real-world depression recognition systems.

\begin{table}[t]
\centering
\caption{Class Distribution for MPDD Dataset}
\vspace{-0.2cm}
\label{tab:data_distribution}
\begin{adjustbox}{max width=\linewidth}
\begin{tabular}{llllll}
\toprule
Task & Label & \multicolumn{2}{c}{Elderly} & \multicolumn{2}{c}{Young} \\
\cmidrule(lr){3-4} \cmidrule(lr){5-6}
 & & \#Samples (Ratio) & \#Spk & \#Samples (Ratio) & \#Spk \\
\midrule
\multirow{3}{*}{\rotatebox[origin=c]{90}{Binary}} 
    & Normal         & 258 (76.6\%) & 68 & 135 (51.1\%) & 45 \\
    & Depressed      & 79 (23.4\%)  & 21 & 129 (48.9\%) & 43 \\
    & Total          & 337          & 89 & 264          & 88 \\
\midrule
\multirow{4}{*}{\rotatebox[origin=c]{90}{Trinary}} 
    & Normal         & 138 (40.9\%) & 37 & 135 (51.1\%) & 45 \\
    & Mild           & 120 (35.6\%) & 31 & 99 (37.5\%)  & 33 \\
    & Severe         & 79 (23.4\%)  & 21 & 30 (11.4\%)  & 10 \\
    & Total          & 337          & 89 & 264          & 88 \\
\midrule
\multirow{6}{*}{\rotatebox[origin=c]{90}{Quinary}} 
    & Normal         & 235 (69.7\%) & 62 & - & - \\
    & Mild           & 68 (20.2\%)  & 18 & - & - \\
    & Moderate       & 23 (6.8\%)   & 6  & - & - \\
    & Severe         & 8 (2.4\%)    & 2  & - & - \\
    & Very Severe    & 3 (0.9\%)    & 1  & - & - \\
    & Total          & 337          & 89 & - & - \\
\bottomrule
\end{tabular}
\end{adjustbox}
\vspace{-4mm}
\end{table}

\section{The first Multimodal Personality-aware Depression Detection Challenge}
This section provides an overview of the MPDD setup, including the datasets, available audio, visual, and text features, as well as the baseline system, which will serve as the foundation for this study.

\subsection{Datasets}
The MPDD dataset comprises two tracks corresponding to distinct age groups, MPDD-Elderly and MPDD-Young, designed to facilitate age-specific depression analysis. Table~\ref{tab:data_distribution} provides a comprehensive summary of class distributions across three classification tasks (binary, trinary, and quinary) for both subsets, reporting sample counts, class ratios, and the number of unique speakers (patients)\footnote{Note that the table only lists the statistics of the MPDD training set. As the time we are writing, the labels of test set are not available. In the following experiments, we split the training set into a 90-10 ratio, using 10\% as the development set and reporting the results based on this split.}.

\subsubsection{Track 1: MPDD-Elderly}

Track 1 focuses on depression detection among elderly participants (average age: 62.8 $\pm$ 11.0).  
Data were collected through semi-structured interviews conducted in hospital settings. Each participant completed standardized clinical questionnaires, including the PHQ-9 and HAMD-24 scales \cite{hamilton1986hamilton}, to assess depression severity. HAMD-24 scores are used to generate labels for the binary and ternary classification tasks, while PHQ-9 scores are used for the quinary classification task.

To enable a more comprehensive participant profile, additional annotations are provided, including Big five personality traits (using a 10-point scale) \cite{rammstedt2007measuring}, physical health conditions, financial stress levels, and the number of cohabiting family members, see Table~\ref{tab:feature_modalities}.

\subsubsection{Track 2: MPDD-Young}

Track 2 targets a younger population (average age: 20.0 $\pm$ 2.2), recruited in non-clinical environments.  
The data collection protocol consists of a self-introduction, a questionnaire segment, and a scripted reading task, all recorded via video.  
Participants completed the PHQ-9 questionnaire, which is used to generate depression labels for the binary and ternary classification tasks.

Personality trait annotations in this track differ slightly from those in MPDD-Elderly. In addition to the Big five traits, demographic variables such as age, gender, and place of origin are included, allowing for comparative analysis across different population groups, see Table~\ref{tab:feature_modalities}.

\begin{table}[t]
\centering
\renewcommand{\arraystretch}{1.1} 
\caption{Feature Modalities and Dimensions in MPDD Dataset}
\vspace{-0.2cm}
\label{tab:feature_modalities}
\begin{adjustbox}{max width=\linewidth}
\begin{tabular}{lll}
\toprule
 & \textbf{Feature Type} & \textbf{Dimensions} \\
\midrule
\multirow{3}{*}{\rotatebox[origin=c]{90}{Audio}} 
    & MFCC\footnotemark[1] & 64 \\
    & OpenSMILE\footnotemark[2] & 6,373 \\
    & Wav2Vec2\footnotemark[3] & 512 \\
\midrule
\multirow{3}{*}{\rotatebox[origin=c]{90}{Visual}} 
    & DenseNet\footnotemark[4] & 1,024 (Elderly) / 1,000 (Young) \\
    & ResNet\footnotemark[5] & 1,000 \\
    & OpenFace\footnotemark[6] & 709 \\
\midrule
\multirow{8}{*}{\rotatebox[origin=c]{90}{Text}} 
    & \multicolumn{2}{l}{\textit{Raw personality traits for MPDD-Elderly:}} \\
    &\multicolumn{2}{l}{Big five: extraversion, agreeableness, openness, neuroticism, conscientiousness}   \\
   & \multicolumn{2}{l}{Disease category: healthy, other, endocrine, circulatory, neurologica}  \\
   & \multicolumn{2}{l}{Financial stress: none, mild, moderate, severe/unbearable}\\
   & \multicolumn{2}{l}{Family members: number of cohabiting individuals} \\
       \cline{2-3}

    & \multicolumn{2}{l}{\textit{Raw personality traits for MPDD-Young:}} \\
    &\multicolumn{2}{l}{Big five, Age, Gender, Native place} \\

    \cline{2-3}
    & \multicolumn{2}{l}{\textit{Personalized feature derived from raw personality traits:}} \\
    & RoBERTa-large\footnotemark[7] & 1,024 \\
\bottomrule
\end{tabular}
\end{adjustbox}
\vspace{-4mm}

\end{table}

\footnotetext[1]{\url{https://github.com/librosa/librosa}}
\footnotetext[2]{\url{https://github.com/audeering/opensmile}}
\footnotetext[3]{\url{https://github.com/facebookresearch/fairseq/tree/main/examples/wav2vec}}
\footnotetext[4]{\url{https://github.com/liuzhuang13/DenseNet}}
\footnotetext[5]{\url{https://huggingface.co/microsoft/resnet-50}}
\footnotetext[6]{\url{http://multicomp.cs.cmu.edu/resources/openface/}}
\footnotetext[7]{\url{https://huggingface.co/FacebookAI/roberta-large}}

\subsection{Feature Modalities}
Audio features include Mel-frequency cepstral coefficients (MFCCs), low-level acoustic descriptors extracted using OpenSMILE \cite{eyben2013opensmile}, and deep learning-based representations from pre-trained models such as Wav2Vec 2.0 \cite{baevski2020wav2vec}. These features provide a comprehensive view of both handcrafted and learned paralinguistic cues.
Visual features comprise deep CNN-based facial embeddings obtained using architectures like DenseNet and ResNet \cite{huang2017densely, he2016deep}, as well as facial behavior analysis (e.g., eye gaze, and head pose) extracted using OpenFace \cite{baltrusaitis2018openface}.
For the textual modality, both tracks offer RoBERTa-based embeddings \cite{liu2019roberta} derived from raw personality traits descriptions. 
Each feature type is provided as a fixed-length embedding per 1-second or 5-second hopping window. These variable-length sequences are temporally aligned on a per-subject basis to maintain consistency across modalities.

\begin{figure*}[!t]
    \centering
\includegraphics[width=0.9\textwidth, trim=2pt 2pt 2pt 1pt, clip]{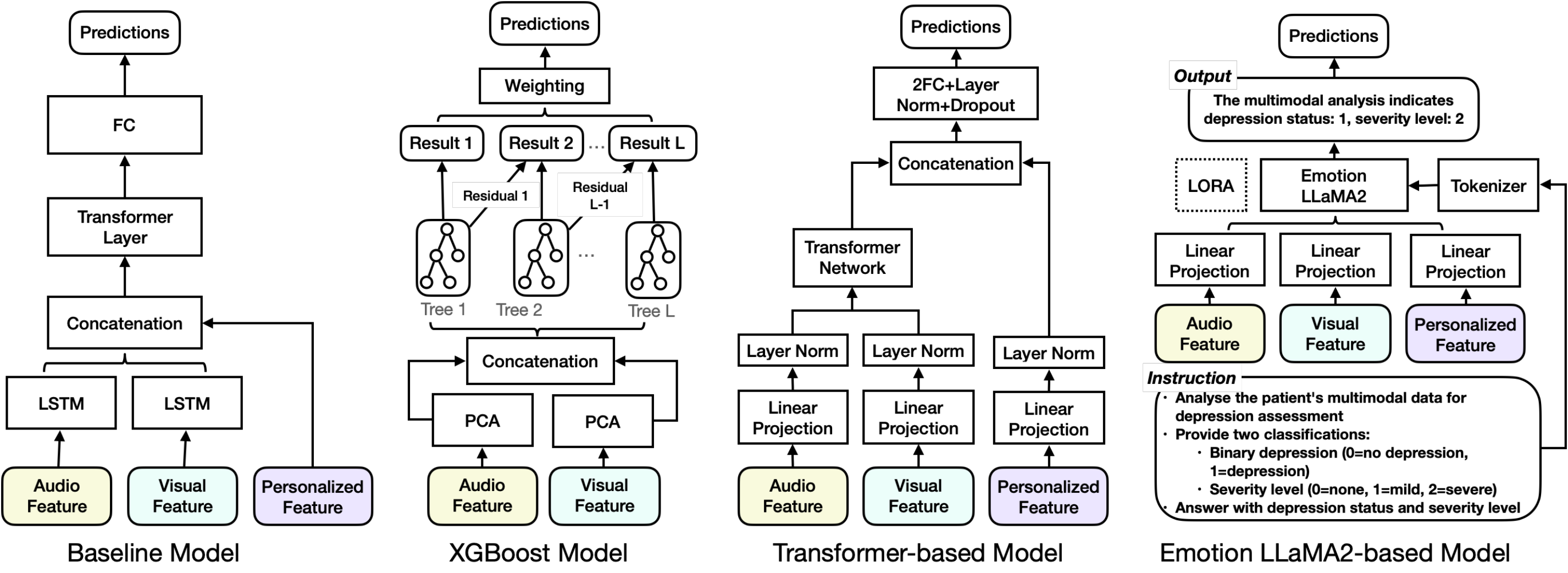}
\caption{Various multimodal depression detection models. The leftmost is the baseline model, while the right three are the models investigated in this paper.}
    \label{fig:overall}
    \vspace{-7mm}
\end{figure*}

\subsection{Baseline System}
The official baseline system adopts a multimodal deep learning approach, as illustrated on the left of Figure~\ref{fig:overall}.
Audio and visual features are first passed through modality-specific encoders, implemented as one-layer LSTM. An optional personalized feature, extracted from a RoBERTa-large model, can be concatenated with the LSTM-processed audio and visual embeddings. The fused representation is then passed through a one-layer transformer that integrates the multimodal information. Finally, the output is fed into fully connected layers to predict the number of classes corresponding to depression severity levels.
The system is trained end-to-end using a combination of cross-entropy loss and focal loss \cite{lin2017focal}, and is evaluated using weighted/unweighted F1 scores and overall accuracy.

\section{Investigated Systems}
This section elaborates on the systems we explored for multimodal depression detection, covering traditional machine learning models, deep learning models, and LLMs.

\subsection{XGBoost-Based Model}

We implemented a gradient-boosted decision tree pipeline using XGBoost for multimodal depression classification based on audio and visual features\footnote{We attempted to incorporate personalized features but did not observe performance improvements; thus, we excluded them from the XGBoost model.}, as illustrated in the second panel of Figure~\ref{fig:overall}. 
Each input sample consists of fixed-length, pre-extracted embeddings derived from pretrained models and represent 1s or 5s window-level summaries of audio and video segments, denoted as $X_a \in \mathbb{R}^{n \times d_a}$ and $X_v \in \mathbb{R}^{n \times d_v}$, respectively, where $n$ is the number of frames for the audio and visual streams, respectively, and $d_a$, $d_v$ are the corresponding feature dimensions. 

To reduce redundancy and improve generalization, we applied Principal Component Analysis (PCA) separately to each modality. For each $X_m \in \{X_a, X_v\}$, we centered the data, computed the covariance matrix $C_m$, and extracted the top-$k$ eigenvectors $V_m^k$. Each modality was then projected as $Z_m = (X_m - \text{mean}(X_m)) V_m^k \in \mathbb{R}^{n \times k}$, with \( k = 50 \). The reduced audio and visual features $Z_a$ and $Z_v$ were concatenated to form a fused multimodal embedding $Z = [Z_a \| Z_v] \in \mathbb{R}^{n \times 100}$, which served as input to the XGBoost classifier.

XGBoost is trained for \( T \) boosting rounds. At each round \( t \), the model computes the gradients \( g_t \) and Hessians \( h_t \) of the multi-class log loss with respect to the current predictions \( \hat{y}^{(t-1)} \), fits a regression tree \( f_t \) to predict the gradients, and updates the predictions as: $\hat{y}^{(t)} \gets \hat{y}^{(t-1)} + f_t(Z)$.
The final model aggregates all trees: $\hat{y}(x) = \sum_{t=1}^{T} f_t(x)$.

Besides XGBoost, to address class imbalance, we apply \textit{class weighting} by assigning a higher weight to the minority class. Specifically, the positive class weight is computed as: 
$w_{\text{pos}} = \frac{N_{\text{neg}}}{N_{\text{pos}}}$, where \(N_{\text{neg}}\) and \(N_{\text{pos}}\) are the number of negative and positive samples, respectively. 
For example, with 90 negatives and 10 positives, \(w_{\text{pos}} = 9\). This helps the model better learn from underrepresented classes.

\subsection{Transformer-Based Model}

We design a multimodal transformer model to fuse audio, visual, and text features for depression classification.

Each input sample \(i\) contains audio \(X_a^{(i)} \in \mathbb{R}^{T_a \times d_a}\), visual \(X_v^{(i)} \in \mathbb{R}^{T_v \times d_v}\), and text features \(x_t^{(i)} \in \mathbb{R}^{d_t}\). These inputs are projected into a shared latent space of dimension \(d\) using learned linear layers followed by layer normalization:
\[
\begin{aligned}
Z_a^{(i)} &= \mathrm{LayerNorm}(X_a^{(i)} W_a + b_a) \in \mathbb{R}^{T_a \times d}, \\
Z_v^{(i)} &= \mathrm{LayerNorm}(X_v^{(i)} W_v + b_v) \in \mathbb{R}^{T_v \times d}, \\
z_t^{(i)} &= \mathrm{LayerNorm}(x_t^{(i)} W_t + b_t) \in \mathbb{R}^{d},
\end{aligned}
\]
where \(W_a \in \mathbb{R}^{d_a \times d}\), \(W_v \in \mathbb{R}^{d_v \times d}\), and \(W_t \in \mathbb{R}^{d_t \times d}\) are learnable projection matrices.
Positional encodings \(P_a \in \mathbb{R}^{T_a \times d}\) and \(P_v \in \mathbb{R}^{T_v \times d}\) are added to preserve temporal order:
\[
\hat{Z}_a^{(i)} = Z_a^{(i)} + P_a, \quad \hat{Z}_v^{(i)} = Z_v^{(i)} + P_v.
\]
The temporally encoded sequences are passed through modality-specific transformer encoders:
\[
\begin{aligned}
H_a^{(i)} &= \mathrm{Transformer}(\hat{Z}_a^{(i)}) \in \mathbb{R}^{T_a \times d}, \\
H_v^{(i)} &= \mathrm{Transformer}(\hat{Z}_v^{(i)}) \in \mathbb{R}^{T_v \times d}.
\end{aligned}
\]
To obtain fixed-length representations, we apply learned attention pooling over time:
\[
\begin{aligned}
\tilde{x}_a^{(i)} &= \sum_{t=1}^{T_a} \alpha_{a,t} H_{a,t}^{(i)} \in \mathbb{R}^d, \\
\tilde{x}_v^{(i)} &= \sum_{t=1}^{T_v} \alpha_{v,t} H_{v,t}^{(i)} \in \mathbb{R}^d.
\end{aligned}
\]
The pooled audio, pooled visual, and text features are concatenated to form the multimodal representation: $z^{(i)} = \left[ \tilde{x}_a^{(i)} \; \| \; \tilde{x}_v^{(i)} \; \| \; z_t^{(i)} \right] \in \mathbb{R}^{3d}.$ Finally, \(z^{(i)}\) is passed through two fully connected layers with ReLU activation, layer normalization, and dropout to predict depression severity. The focal loss function is used to handle class imbalance.

It is well recognized that deep learning models are prone to overfitting when trained on small datasets, as is the case with the MPDD dataset. To address this, we apply \textit{Mixup} data augmentation \cite{zhang2018mixup} during training, which has been shown to improve model robustness and performance across various deep learning tasks.  The mixup augmentation strategy creates synthetic training examples by linearly interpolating between pairs of samples and their corresponding labels. Specifically, a new example is generated by combining the two input samples and labels using a mixing coefficient.

\subsection{LLM-Based Model}
We also explore the use of LLM-based methods for the MPDD task. Specifically, we are inspired by Emotion-LLaMA~\cite{NEURIPS2024_c7f43ada}, a model built upon the LLaMA backbone and fine-tuned on large-scale multimodal emotion datasets. Emotion-LLaMA enables emotion reasoning by integrating visual, auditory, and textual cues through structured prompts and cross-modal attention.

Building upon this foundation, we adapt Emotion-LLaMA to the MPDD task by fine-tuning it on our multimodal dataset. The model formulation is expressed as:
\begin{equation}
O = \phi\big(\sigma_{\text{aud}}(X_a^{(i)}), \sigma_{\text{vis}}(X_v^{(i)}), \sigma_{\text{txt}}(x_t^{(i)}), \text{Tokenizer}(\text{Prompt})\big).
\end{equation}
Here, the input consists of audio features \(X_a^{(i)}\), visual features \(X_v^{(i)}\), textual features \(x_t^{(i)}\), and a task-specific prompt in a multiple-choice question format (as illustrated on the rightmost side of Figure~\ref{fig:overall}). 

To integrate features from multiple modalities, we introduce a linear projection mechanism that maps each modality into a shared embedding space. This is achieved via trainable linear projection functions: \(\sigma_{\text{aud}}\) for audio, \(\sigma_{\text{vis}}\) for visual, and \(\sigma_{\text{txt}}\) for text. The final output \(O\) is a formatted text response, also shown in the rightmost bottom of Figure~\ref{fig:overall}.

The fine-tuning process involves two stages. In Stage 1, the LLaMA backbone is frozen, and only the projection layers and classification heads are trained. In Stage 2, LoRA-based fine-tuning is applied, using a dual learning rate setup to update both the LoRA parameters and the projection layers.

\begin{table}[t]
\caption{Ablation study on each systems for 5s MPDD Elderly binary dev set (90(cross-validation)/10).}
\vspace{-0.2cm}
\label{tab:ablation}
\centering
\begin{tabular}{lcc}
\toprule
\textbf{Methods} &  $W_{F1}$& $U_{F1}$\\
\midrule
\textbf{XGBoost} &  \\
\quad  Raw Feature & 65.76 & 42.62 \\
\quad  Raw Feature + \textit{class weighting} & 74.80 & 65.04 \\
\quad  PCA Feature + \textit{class weighting} & 94.29 & 91.90 \\

\midrule

\textbf{Transformer} &  \\
\quad 2 transformer layers &  80.44 & 69.64\\
\quad 2 transformer layers  + \textit{mixup} &  84.00 & 76.22\\
\quad 2 transformer  layers  +\textit{mixup}  + \textit{cross-validation} &  87.08 & 83.19 \\

\midrule

\textbf{LLM} &  \\
\quad Llama2 &  60.96 & 44.57\\
\quad EmotionLlama2 + 1step &  31.77 & 33.33\\
\quad EmotionLlama2 + 2steps &  70.59 & 52.55\\

\bottomrule
\end{tabular}
\vspace{-5mm}
\end{table}

\begin{table*}[]
\centering
\caption{Weighted F1 and Unweighted F1 (\%) $\uparrow$ results on MPDD-Elderly dev set.}
\vspace{-0.2cm}
\begin{tabular}{l|c|cc|cc|cc|cc|cc|cc}
\toprule

\multirow{3}{*}{Method}  & \multirow{3}{*}{PF} & \multicolumn{6}{c|}{1s} & \multicolumn{6}{c}{5s} \\

& 
& \multicolumn{2}{c}{Binary} 
& \multicolumn{2}{c}{Ternary} 
& \multicolumn{2}{c|}{Quinary}

& \multicolumn{2}{c}{Binary} 
& \multicolumn{2}{c}{Ternary} 
& \multicolumn{2}{c}{Quinary} 
 \\
 & &  $W_{F1}$& $U_{F1}$& $W_{F1}$& $U_{F1}$& $W_{F1}$& $U_{F1}$ & $W_{F1}$& $U_{F1}$& $W_{F1}$& $U_{F1}$& $W_{F1}$& $U_{F1}$\\
\midrule
Baseline  & \xmark  
& 82.60 & 70.89 &54.35 & 49.14 & 63.85 & 44.00
& 77.90 & 66.15 &50.88 & 47.59 & 73.49 & 56.83 \\
Baseline & \cmark 
& 85.71 & 79.13 &56.48 & 55.64 & 66.26 & 46.66 
& 81.75 & 72.37 &58.22 & 59.37 & 75.62 & 58.40\\
\midrule
XGBoost   & \xmark   
& 90.67 & 85.83 &55.23 & 53.02 & 55.60 & 21.43 
& \textbf{94.29} & \textbf{91.90} & 61.02 & \textbf{62.51} & 54.62 & 21.05\\

Transformer & \cmark   
& \textbf{93.44} & \textbf{88.21} & \textbf{74.95} & \textbf{80.00} & \textbf{82.21} & \textbf{46.77} 
& 85.27 & 71.55 &\textbf{65.52} & 61.31 & 67.96 & \textbf{67.62} \\

LLM  & \cmark  
& 70.59 & 52.55 &61.17 & 53.02 & 77.89 & 30.60
& 67.43 & 45.60 &45.66 & 37.44 & {\textbf{77.89}} & {30.60} \\

\bottomrule
\end{tabular}
\label{tab:result-elder}
\end{table*}

\begin{table*}[]
\centering
\caption{Weighted F1 and Unweighted F1 (\%) $\uparrow$ results on MPDD-Young dev set with model size for the binary task.}
\vspace{-0.2cm}
\begin{tabular}{l|c|c|cc|cc|cc|cc}
\toprule

\multirow{3}{*}{Method}  & \multirow{3}{*}{PF} & \multirow{3}{*}{\#Params(M)} & \multicolumn{4}{c|}{1s} & \multicolumn{4}{c}{5s} \\

& & 
& \multicolumn{2}{c}{Binary} 
& \multicolumn{2}{c|}{Ternary} 
& \multicolumn{2}{c}{Binary} 
& \multicolumn{2}{c}{Ternary} 
 \\
 & & &  $W_{F1}$& $U_{F1}$& $W_{F1}$& $U_{F1}$& $W_{F1}$& $U_{F1}$ & $W_{F1}$& $U_{F1}$ \\
\midrule
Baseline  & \xmark & 1.89\footnotemark[1]
& 55.23 & 55.23 &47.95 & 43.72 
& 60.02 & 60.02 &42.82 & 39.38  \\
Baseline & \cmark & 2.15
& 59.96 & 59.96 &51.86 & 51.62  
& 62.11 & 62.11 &48.18 & 41.31 \\
\midrule
XGBoost   & \cmark & 0.002
& 81.53 & 81.38 &66.67 & 48.89 
& 74.07 & 74.07 & 62.19 & 45.60\\

Transformer & \cmark & 1.06
& \textbf{95.83} & \textbf{95.83} & \textbf{75.60} & \textbf{71.36}  
& \textbf{81.48} & \textbf{81.48} &\textbf{78.51} &\textbf{59.16}  \\

LLM  & \cmark & 6,843
& 60.86 & 61.65 &45.65 & 34.06 
& 64.07 & 64.96 &39.49 & 28.82 \\

\bottomrule
\end{tabular}
\label{tab:result-young}
\vspace{-0.2cm}
\end{table*}
\footnotetext[1]{The parameters of the baseline system are calculated using wav2vec and OpenFace features as the input features.}

\section{Experiments}
Experiments begin with the 5-second MPDD-Elderly dataset, where we conduct a comprehensive ablation study to identify the optimal configuration for each system. Once the best settings are determined, these configurations are applied to the remaining scenarios. For all scenarios, we use a 90-10 train-validation split based on patient IDs, ensuring subject independence and preventing information leakage.
We evaluated each configuration using two primary metrics: weighted ($W_{F1}$) and unweighted F1 scores ($U_{F1}$).

\subsection{Experiment Settings}
\subsubsection{XGBoost Setting}

The XGBoost configuration was carefully designed to balance interpretability, computational efficiency, and robust performance on limited multimodal data. 
Each audio and visual modality was first reduced to 50 dimensions using PCA, resulting in a fused 100-dimensional feature vector. 
For configurations without PCA, original modality features were concatenated directly. 
The model used a shallow tree depth with a maximum depth of 3, a learning rate selected from \(\{0.01, 0.05\}\), and both subsample and colsample-by-tree ratios set to 0.8 to introduce randomness and reduce overfitting. Training employed up to 500 boosting rounds with early stopping after 25 rounds without improvement, using multi-class log loss as the evaluation metric and \texttt{multi:softprob} as the objective function. Hyperparameters were tuned per modality pair and classification task (binary, ternary, quinary) using early stopping on a speaker-level validation split to avoid overfitting and data leakage. Multiple audio-visual fusion pairs were explored (e.g., MFCC + OpenFace, OpenSMILE + ResNet), with MFCC + OpenFace achieving the best results. Consistency was maintained by applying the same patient-level ID split and evaluation methodology across all experiments. Notably, personality-aware features and probabilistic model ensembling were excluded to enable a focused evaluation of core modality fusion performance under classical machine learning settings.

\subsubsection{Transformer setting}
The transformer configuration prioritizes efficiency and regularization to avoid overfitting on the relatively small dataset. The model uses a reduced dimensionality of 128, with a shallow 2-layer transformer architecture and 4 attention heads, which together provide sufficient representational capacity while maintaining computational efficiency. A dropout rate of 0.5 is applied throughout the network for strong regularization. The pre-classifier network reduces the concatenated multimodal embedding from 512 to 256 dimensions, refining the joint feature representation for downstream classification.
For training, we use a learning rate of 5e-5, batch size of 2, and train for up to 100 epochs, with early stopping triggered if no improvement is observed for 20 consecutive epochs. We also include warmup training for the first 10 epochs, gradient clipping at 1.0, and apply weight decay of 1e-4 to further aid generalization. 
Multiple audio-visual fusion pairs were explored, and wav2vec2, DenseNet, with personalized features were selected.

For the mixup augmentation strategy, the mixing coefficient is sampled from a Beta distribution with parameters 0.2 and 0.2. Applied consistently across all modalities with a 50\% probability, mixup helps the model learn smoother decision boundaries and improves generalization on limited data. The cross-validation experiment employs 10-fold validation, where each fold naturally provides the 90-10 split used in strategies without cross-validation.

\subsubsection{LLM setting}

The Depression-LLaMA implementation is based on the Emotion-LLaMA pre-trained foundation, utilizing the LLaMA-2-7B\footnote{\url{https://huggingface.co/meta-llama/Llama-2-7b}} as the underlying large language model. 
The model architecture utilize all features, including three audio features, three visual feature and one text features listed in table \ref{tab:feature_modalities}
Each of them will be mapped to 4,096 dimensional features by the linear projection. The training process is conducted in two stages. In Stage 1, the backbone is frozen and trained for 5 epochs with a learning rate of \(5 \times 10^{-5}\). In Stage 2, LoRA fine-tuning is applied for 3 epochs with a learning rate of \(1 \times 10^{-5}\).

\subsection{Results}
\subsubsection{Ablation Study for each system}

Table~\ref{tab:ablation} presents the weighted F1 ($W_{F1}$) and unweighted F1 ($U_{F1}$) scores for various systems evaluated on the 5-second MPDD Elderly binary development set using a 90-10 split (with cross-validation where specified). The results compare baseline methods, traditional machine learning approaches, transformer models, and LLM baselines,  with different training strategies.

\noindent
\textbf{XGBoost}:
Starting with raw features, XGBoost achieves moderate performance. Incorporating class weighting leads to a notable improvement, and applying PCA alongside class weighting further enhances results, demonstrating the effectiveness of dimensionality reduction and handling class imbalance.

\noindent
\textbf{Transformer}:
The baseline transformer outperforms XGBoost without PCA, showing strong capability in modeling the data. Adding mixup augmentation further improves the model’s generalization, and combining mixup with cross-validation yields the best and most robust performance, highlighting the benefits of data augmentation and rigorous evaluation.

\noindent
\textbf{LLM Methods}:
The baseline LLaMA2 model performs relatively poorly compared to other methods. Initial fine-tuning of EmotionLLaMA2 results in a performance drop, likely due to adaptation challenges, but subsequent fine-tuning improves outcomes considerably. Despite this, LLM-based methods still lag behind the transformer and XGBoost models on this task.

\subsubsection{Overall Results}

Table~\ref{tab:result-elder} and Table~\ref{tab:result-young} present a comprehensive comparison of all explored methods on the MPDD-Elderly dataset.
On the MPDD-Elderly development set, XGBoost achieves the highest weighted and unweighted F1 scores for the 5-second binary classification task, demonstrating strong performance on longer audio segments with effective feature engineering. The Transformer model outperforms XGBoost on shorter 1-second segments and more complex classification tasks (ternary and quinary), indicating its strength in capturing fine-grained information. Personalized features improve baseline models but the LLM-based approach shows comparatively lower performance across most tasks.

For the MPDD-Young development set, the Transformer consistently delivers the best results across all classification tasks and time windows, with weighted F1 scores reaching as high as 95.83\% on 1-second binary classification. XGBoost performs well but lags behind Transformer models, especially on ternary tasks. Baseline models benefit from personalized features but remain less competitive, while LLM-based methods perform the weakest.

Overall, the Transformer model (1.06M parameters) is the most effective for multimodal depression detection, especially for younger speakers and shorter audio windows. Despite having significantly more parameters, LLM (6,843M) underperforms, especially in ternary classification, suggesting that larger models don’t always guarantee better performance. XGBoost, with only 0.002M parameters, achieves strong results in binary classification, highlighting the effectiveness of simpler models for specific tasks. The Baseline with a personalized feature slightly improves on the baseline without it but remains less effective than more complex models, even though it has more parameters (2.15M) than the Transformer.

\noindent
\textbf{Acknowledgment}
This research is supported by the Ministry of Education, Singapore, under its Academic Research Tier 1 (Grant number: GMS 956) and the Academy of Medical Sciences, under its Networking Grant (NGR1\textbackslash1678). 

\bibliographystyle{IEEEtran}
\bibliography{mybib}
\end{document}